\documentclass[conference]{IEEEtran}
\IEEEoverridecommandlockouts
\usepackage{cite}
\usepackage{amsmath,amssymb,amsfonts}
\usepackage{algorithmic}
\usepackage{graphicx}
\usepackage{textcomp}
\usepackage{xcolor}
\def\BibTeX{{\rm B\kern-.05em{\sc i\kern-.025em b}\kern-.08em
    T\kern-.1667em\lower.7ex\hbox{E}\kern-.125emX}}
\begin{document}

\title{Block Pruning for Enhanced Efficiency in Convolutional Neural Networks
\\
\thanks{}
}

\author{\IEEEauthorblockN{ Cheng-En Wu}
\IEEEauthorblockA{\textit{University of Wisconsin-Madison} \\
cwu356@wisc.edu}
\and
\IEEEauthorblockN{Azadeh Davoodi}
\IEEEauthorblockA{\textit{University of Wisconsin-Madison} \\
adavoodi@wisc.edu}
\and
\IEEEauthorblockN{Yu Hen Hu}
\IEEEauthorblockA{\textit{University of Wisconsin-Madison} \\
yhhu@wisc.edu}
% \and
% \and
% \and
% \and
}

\maketitle

\begin{abstract}
This paper presents a novel approach to network pruning, targeting block pruning in deep neural networks for edge computing environments. Our method diverges from traditional techniques that utilize proxy metrics, instead employing a direct block removal strategy to assess the impact on classification accuracy. This hands-on approach allows for an accurate evaluation of each block's importance. We conducted extensive experiments on CIFAR-10, CIFAR-100, and ImageNet datasets using ResNet architectures. Our results demonstrate the efficacy of our method, particularly on large-scale datasets like ImageNet with ResNet50, where it excelled in reducing model size while retaining high accuracy, even when pruning a significant portion of the network. The findings underscore our method's capability in maintaining an optimal balance between model size and performance, especially in resource-constrained edge computing scenarios. 
\end{abstract}


%%%%%%%%% BODY TEXT
% %--------------------------Introduction----------------------------------
\section{Introduction}
\label{sec:intro}
In the rapidly evolving landscape of edge computing, the deployment of machine learning models on edge devices is constrained by limited memory and computational power. To address these challenges, we present an innovative approach that leverages pre-existing pre-trained models as an initiation point. Our primary objective is to derive pruned models that demand less computational power, thus yielding quicker inference speeds and reduced memory consumption.

We commence with block pruning, and depending on the necessity, progress to channel pruning followed by weight pruning. After each pruning phase, we harness the power of transfer learning to fine-tune the modified units. Recognizing the inefficiencies of exhaustive search techniques, we introduce heuristic methods to identify potential candidates for pruned models. Our proposed heuristic methods not only economize on time but also closely approximate optimal solutions.

Block pruning for convolutional models includes the accurate estimation of a model's performance after selective block removal. Brute force search methods have relied on exhaustive search, which is computationally expensive and falls within the NP-hard problem space—a terrain not favorable to computational efficiency. To this end, the question arises: can we employ heuristic polynomial-time solutions to tackle this subset selection problem efficiently?

Existing algorithms\cite{chen2018shallowing,Elkerdawy_2020_ACCV,tang2023srinit,zhang2022layer_shallower}, offer varying approaches to this issue, generally adopting proxy methods for performance estimation. These methods only provide an approximate inference of the model's performance based on the perturbation of selected subsets, often relying solely on validation data for evaluation. Such an approach inevitably results in less accurate performance metrics and leaves unaddressed challenges like dimensionality mismatch following block removal.

In contrast, our proposed method adopts a more direct approach by executing block pruning first and then applying transfer learning to fine-tune the pruned architecture. This allows for an exact evaluation of the model's performance, as opposed to relying on proxy methods. Our approach also accounts for potential issues like dimensionality mismatch by incorporating additional transfer learning steps. The result is a more accurate, and potentially more efficient, model pruning technique.

% %--------------------------Related Work----------------------------------
\section{Related Work}
\label{sec:r_work}

Block pruning in deep neural networks (DNNs) is essential for optimizing computational efficiency and reducing storage requirements, particularly in resource-constrained environments. Various methods have been developed, each addressing this challenge uniquely. Sparse Structure Selection (SSS)\cite{huang2018data} integrates pruning and training through an L1 regularization approach, focusing on scaling factors to align the pruning process with ongoing training, thereby ensuring minimal impact on network performance while enhancing efficiency. Proxy Classifier-based Methods, such as those proposed by Chen et al.\cite{chen2018shallowing} and Elkerdawy et al.\cite{Elkerdawy_2020_ACCV}, utilize proxy classifiers for block selection based on intermediate feature representations, pruning blocks with minimal impact on overall performance. Discrimination Based Block-Level Pruning (DBP), proposed by Wang et al.\cite{wang2019dbp}, accelerates deep model performance by targeting blocks that contribute less to the model's discrimination capacity. SRinit\cite{tang2023srinit} employs stochastic re-initialization of layer parameters to identify layers that are less sensitive to such changes. Layers with low accuracy drop upon re-initialization are deemed less crucial and suitable for pruning, thereby maintaining the model's performance integrity.

% %--------------------------Methodology-----------------------------------
\section{Proposed Method}
\label{sec:method}

\subsection{Types of Convolutional blocks}
Convolutional blocks are integral components of many modern deep learning architectures. They encompass a variety of structural designs, each with specific functions and characteristics. In the following, we describe common types of convolutional blocks, highlighting their key features and roles in enhancing network performance and efficiency. 

\noindent \textbf{Fully Connected (FC) block:} This block is composed of a fully connected (FC) layer coupled with a ReLu (Rectified Linear Unit) layer. The FC layer offers global context by connecting every neuron in the layer to every neuron in the preceding layer. The ReLu layer is a non-linear operation that helps the model learn complex patterns within the data.

\noindent \textbf{Basic Block:} The basic block is made up of a fully Convolutional (Conv) layer and a ReLu layer. The Conv layer performs a dot product between its weights and a small region it is connected to in the input volume, thereby preserving the spatial relationship between pixels.

\noindent \textbf{Pooling Block:} This block is constructed with a fully Convolutional (Conv) layer, followed by a Pooling and ReLu layer. The pooling operation reduces the spatial size of the Conv layer, thereby reducing the computational complexity and the number of parameters, and in turn, the risk of overfitting.

\noindent \textbf{Residual Block\cite{he2016deep}:} This block contains a series of Conv layers and ReLu activations followed by a shortcut or skip connection. The Residual block enables the training of deep networks by allowing the gradient to be directly backpropagated to earlier layers.

\noindent \textbf{Residual Block w/ 1x1 Conv:} Similar to the standard Residual block, but with the addition of a 1x1 Conv layer. The 1x1 Conv layer, also known as pointwise convolution, changes the dimensionality of the input feature maps either by increasing or reducing the number of channels, adding an extra level of abstraction.

\noindent \textbf{Dense Block\cite{huang2017densely}:} Originating from the DenseNet architecture, the Dense block is assembled using multiple convolution blocks, all of which maintain a consistent number of output channels. This structure follows a feed-forward design wherein each layer is directly interconnected with all subsequent layers, thereby promoting feature reuse throughout the network.

\noindent \textbf{MobileNet Block:} This block is based on the MobileNet\cite{howard2017mobilenets} architecture which uses depthwise separable convolutions instead of the standard convolutions. This technique reduces the number of parameters, thus decreasing the computational cost and making the network lighter and faster, which is ideal for mobile and embedded vision applications.

\noindent \textbf{MobileNet V2 Stride=1 Block:} This block is a variation of the MobileNet block that maintains the resolution of the input. It uses inverted residuals and linear bottlenecks to improve the network's efficiency, as introduced in MobileNetV2\cite{sandler2018mobilenetv2}.

\noindent \textbf{MobileNet V2 Stride=2 Block:} Similar to the MobileNet V2 Stride=1 block but with a stride of 2, which means that the block reduces the spatial dimensions of the output by half, further increasing the computational efficiency.

\noindent \textbf{Inception Block:} This block introduced in the Inception network\cite{szegedy2017inception} (or GoogLeNet), consists of parallel paths with convolutions of different sizes (1x1, 3x3, and 5x5), all of which are processed simultaneously. A unique characteristic of the Inception block is that it also includes a parallel max pooling path.

\begin{figure*}[t]
  \begin{center}
    \includegraphics[width=0.9\textwidth]{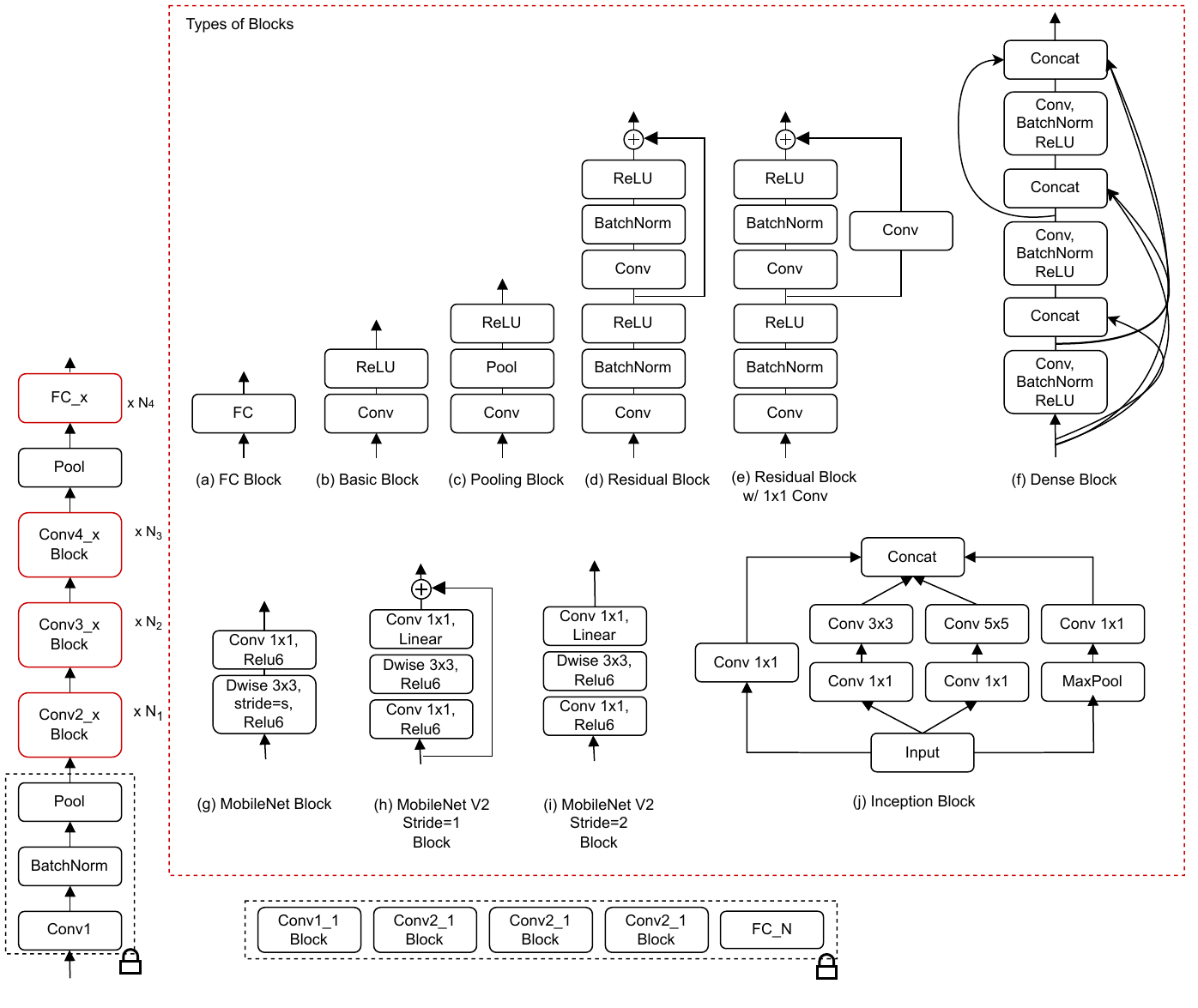}
  \end{center}
  \caption{The diagram shows a deep neural network composed of several stacked Convolutional (Conv) blocks, each of which can be classified into 10 distinct types. We have enforced a limitation that restricts pruning on the truncated Conv block (which is the initial Conv in the network), each of the first blocks within the Conv Blocks, as well as the final fully connected (FC) layer.}
  \label{fig:Types_of_Blocks}
\end{figure*}

\subsection{Baseline}

We use two baseline methods for block pruning in our approach. The first method is sequential block pruning, which prunes blocks from the end of the network to the beginning. This method is based on the principle that later blocks in deep neural networks might be less crucial for maintaining overall network performance, especially in models with a significant depth. The blocks are pruned sequentially, which allows for a systematic reduction in network complexity while monitoring the impact on performance.
The second method is SRinit, as described in \cite{tang2023srinit}. This method adopts a novel approach by adding noise to the blocks to evaluate the importance of each block. Specifically, it involves the stochastic re-initialization of layer parameters, where layers with a low accuracy drop after re-initialization are considered to be more redundant. These layers are less likely to compromise the prediction ability of the network if removed. 

\subsection{Our Method}
Unlike baseline methods that rely on proxy metrics to evaluate the importance of each block, our approach directly removes individual valid blocks to ascertain their impact on classification accuracy. This hands-on strategy offers a genuine reflection of the importance of each pruned block. We identify valid blocks, which do not serve as bridging elements between blocks with different channel numbers, for pruning. As illustrated in Fig\ref{fig:Types_of_Blocks}, certain blocks, like the stem convolutional (Conv) layer and the first block in each stacked layer, are exempted from pruning. Let \( N \) represent our neural network architecture, consisting of a sequence of blocks \( B_1, B_2, \ldots, B_n \), and \( V \) be the set of indices of valid blocks. Our pruning process \( P(N, i) \) involves removing block \( B_i \) from \( N \), where \( i \in V \). The impact of pruning on network performance is measured by the classification accuracy \( A(N') \) of the pruned network. This method not only simplifies the pruning process by addressing the dimension mismatch issue without needing compensatory blocks but also maintains the integrity and performance of the network architecture.

% %--------------------------Experiments-----------------------------------
\section{Experiments}
\label{sec:exp}

\begin{figure}[t]
  \begin{center}
    \includegraphics[width=\columnwidth]{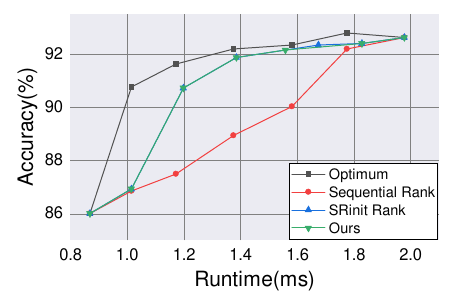}
  \end{center}
  \caption{ResNet20 on Cifar10}
  \label{fig:resnet20_Cifar10_comp}
\end{figure}

\begin{figure}[t]
  \begin{center}
    \includegraphics[width=\columnwidth]{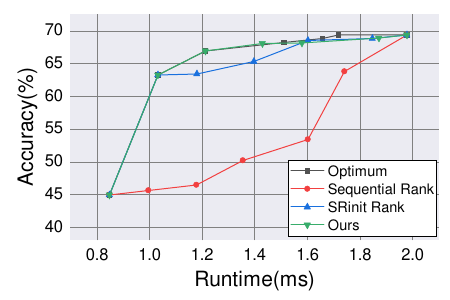}
  \end{center}
  \caption{ResNet20 on Cifar100}
  \label{fig:resnet20_cifar100_comp}
\end{figure}

\begin{figure}[t]
  \begin{center}
    \includegraphics[width=\columnwidth]{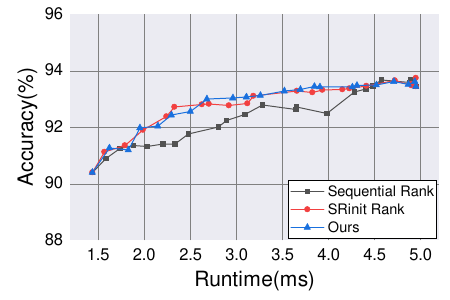}
  \end{center}
  \caption{ResNet56 on Cifar10}
  \label{fig:resnet56_cifar10_comp}
\end{figure}

\begin{figure}[t]
  \begin{center}
    \includegraphics[width=\columnwidth]{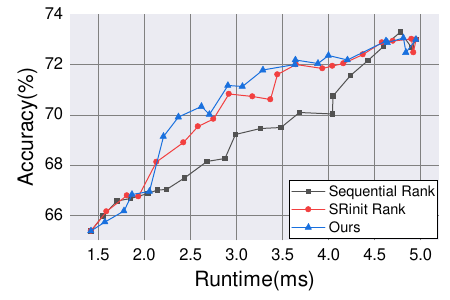}
  \end{center}
  \caption{ResNet56 Cifar100}
  \label{fig:resnet56_cifar100_comp}
\end{figure}

\begin{figure}[t]
  \begin{center}
    \includegraphics[width=\columnwidth]{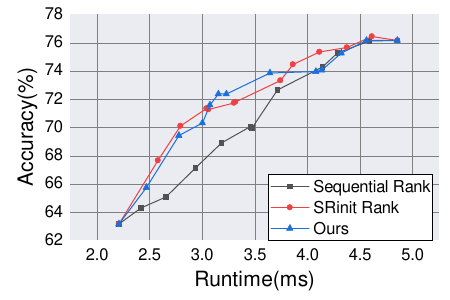}
  \end{center}
  \caption{ResNet50 ImageNet}
  \label{fig:resnet50_imagenet_comp.pdf}
\end{figure}

\subsection{Experimental Setup}

\noindent \textbf{Datasets.} In this study, we primarily focus on three prevalent datasets to validate the efficiency of our method: CIFAR-10, CIFAR-100, and ImageNet. CIFAR-10 and CIFAR-100 are known to consist of 60K 32 × 32 RGB images, with CIFAR-10 segregating them into 10 classes and CIFAR-100 into 100 fine-grained classes. ImageNet, on the other hand, comprises larger 224 × 224 RGB images with 1.2M for training and 50K for testing across 1000 distinct object classes.

\noindent \textbf{Implementation Details.} 
Our implementation involves using ResNet56 for CIFAR-10 and CIFAR-100 datasets, and ResNet50 for the ImageNet dataset. We employ Stochastic Gradient Descent (SGD) as the optimization algorithm across all experiments. To test inference speed, we input a single image to the model and execute the process 1000 times, calculating the average time to evaluate the model's performance post-pruning. 
Following the block pruning process, we fine-tune the networks for further optimization. For CIFAR-10 and CIFAR-100, this fine-tuning phase extends for 50 epochs, starting with an initial learning rate of 0.001. For the ImageNet dataset, we fine-tune for 10 epochs with an initial learning rate set at 0.0001. In all cases, we maintain a consistent weight decay of 0.005 and a batch size of 128. Additionally, a decay factor of 0.5 is applied to the learning rate at specific epochs: at 20, 30, and 40 for CIFAR-10 and CIFAR-100, and at 5 and 8 for ImageNet.

\subsection{Main Results}
In our study, we present an analysis of accuracy versus inference speed for all experiments conducted. The data points commence with the original model, represented by the rightmost point, which includes all blocks. Subsequently, we incrementally remove one block at a time, progressing towards the left in our Figs. This process continues until all valid blocks are pruned.

\noindent \textbf{Comparison with Optimum} 
In our analysis, we focus on ResNet20 due to its manageable size, featuring only six valid blocks eligible for pruning according to our criteria. This allows for a comprehensive brute-force search to identify the best combination of blocks for pruning. We compare our method's performance against two baseline methods. The results, illustrated in Fig\ref{fig:resnet20_Cifar10_comp}, show that our approach outperforms the others on the CIFAR-10 dataset. On CIFAR-100, our method significantly surpasses the baseline methods as shown in Fig\ref{fig:resnet20_cifar100_comp}, indicating its robustness across different datasets. These findings suggest that our pruning strategy yields solutions close to the optimal configuration, demonstrating the efficacy of our approach in efficiently pruning networks while maintaining high performance.

\noindent \textbf{Deeper Network} 
Exploring the applicability of our pruning method to more complex architectures, we extended our experiments to deeper networks. The results from these tests are shown in Fig\ref{fig:resnet56_cifar10_comp} and Fig\ref{fig:resnet56_cifar100_comp}. Despite the increased complexity and depth of these networks, our pruning method consistently outperformed the baseline methods. This was evident across various metrics, most notably in maintaining high accuracy while significantly improving inference speed. The resilience of our approach in handling deeper network structures underscores its robustness and adaptability. These results reinforce the effectiveness of our pruning strategy, even as network complexity scales up, offering a promising solution for optimizing larger, more intricate models in real-world applications.

\noindent \textbf{Large Scale Dataset}
To evaluate the scalability of our approach with large-scale datasets, we applied our pruning method to ResNet50 trained on the ImageNet dataset. The results of this experiment, as depicted in Fig\ref{fig:resnet50_imagenet_comp.pdf}, clearly demonstrate the effectiveness of our method in a more demanding context. Our approach achieved superior results compared to the baseline methods, particularly when the number of pruned blocks was less than half of the total blocks in the network. This indicates not only the efficiency of our pruning technique in reducing the model size but also its ability to retain high accuracy, even with significant reduction in network complexity. These findings highlight the potential of our method in handling large-scale datasets and complex network architectures, providing an optimal balance between model size and performance.

% %--------------------------Conclusion---------------------------------
\section{Conclusion}
\label{sec:conclusion}
In this paper, we introduced a novel block pruning method for deep neural networks, particularly tailored for edge computing environments. Our approach, distinct from traditional pruning methods, directly removes blocks to assess their impact on classification accuracy. This strategy offers a more precise understanding of each block's contribution to the overall network performance. Our extensive experiments with ResNet architectures on CIFAR-10, CIFAR-100, and ImageNet datasets have demonstrated the effectiveness of our method. Notably, on large-scale datasets like ImageNet, our approach achieved significant model size reduction while maintaining high accuracy, even with substantial network pruning. This underscores our method's efficacy in balancing model size and performance, a crucial factor in resource-constrained scenarios.

\section*{Acknowledgement}
\noindent This work was partially supported by the National Science Foundation under Grant No. 2006394.

\clearpage
%%%%%%%%% REFERENCES
{\small
\bibliographystyle{IEEEtran}
\bibliography{IEEEfull}
}

\end{document}